\documentclass[11pt]{article}

\usepackage[final]{acl} 

\usepackage{times}
\usepackage{latexsym}

\usepackage[T1]{fontenc}

\usepackage[utf8]{inputenc}

\usepackage{microtype}

\usepackage{inconsolata}

\usepackage{graphicx}

\usepackage{array}
\usepackage{xspace}
\usepackage{xcolor} 
\usepackage{booktabs}
\usepackage{adjustbox}
\usepackage{subcaption} 
\usepackage{multirow}
\usepackage{makecell}
\usepackage{pifont}
\usepackage{amsmath}
\usepackage{amssymb}
\usepackage{algorithm}
\usepackage{algpseudocode}
\usepackage[inline]{enumitem} 
\setlist[itemize]{leftmargin=*, nosep}
\setlist[enumerate]{leftmargin=*,nosep,label=(\arabic*)}
\usepackage{listings}
\usepackage{float}
\usepackage{longtable}
\usepackage[most]{tcolorbox}
\usepackage{siunitx} 
\usepackage{placeins}

\tcbuselibrary{listings,skins,breakable}

\newcommand{\cmark}{\ding{51}}
\newcommand{\xmark}{\ding{55}}
\newcommand{\pmark}{(\ding{51})}

\newcommand{\ymark}{\makebox[1.0em][l]{\cmark}}  
\newcommand{\nmark}{\makebox[1.0em][l]{\xmark}}

\newcommand{\OurModel}[1]{StoryMI}

\usepackage{acronym}
\acrodef{AI}{artificial intelligence}
\acrodef{LLM}{large language model}
\acrodef{MI}{motivational interviewing}
\acrodef{BC}{behavioral coding}
\acrodef{MISC}{Motivational Interviewing Skill Code}

\definecolor{promptbg}{RGB}{248,249,250}
\definecolor{dialoguebg}{RGB}{253,253,253}
\definecolor{promptframe}{RGB}{200,200,200}
\definecolor{systembg}{RGB}{240,248,255}
\definecolor{clientbg}{RGB}{240,255,240}
\definecolor{therapistbg}{RGB}{255,248,240}
\definecolor{badcolor}{RGB}{220,53,69}
\definecolor{goodcolor}{RGB}{40,167,69}

\newtcolorbox{promptbox}[1][]{
    enhanced,
    breakable,
    colback=white,
    colframe=black!70,
    boxrule=0.5pt,
    left=6pt, right=6pt, top=6pt, bottom=6pt,
    before skip=6pt,
    after skip=6pt,
    fontupper=\small,
    #1
}

\newtcolorbox{dialoguebox}[2][]{
    enhanced,
    breakable,
    colback=#2!3,
    colframe=#2,
    boxrule=1pt,
    left=8pt, right=8pt, top=8pt, bottom=8pt,
    before skip=8pt,
    after skip=8pt,
    fontupper=\small,
    #1
}
\title{StoryMI: Steerable Multi-Agent Therapeutic Dialogue Generation}

\author{
  \textbf{Qingyu Meng\textsuperscript{1}},
  \textbf{Min Chen\textsuperscript{1}},
  \textbf{Dingming Liu\textsuperscript{1,2}},
  \textbf{Yifan Mo\textsuperscript{1}},
\\
  \textbf{Yue Su\textsuperscript{1}},
  \textbf{Xin Sun\textsuperscript{3}},
  \textbf{Koen Hindriks\textsuperscript{1}},
  \textbf{Jiahuan Pei\textsuperscript{1}\thanks{Corresponding author.}}
\\
  \textsuperscript{1}Vrije Universiteit Amsterdam,
  \textsuperscript{2}Bol.com,
  \textsuperscript{3}NII, Tokyo Institute of Technology
\\
  \small{\href{mailto:q.meng@vu.nl}{q.meng@vu.nl}, \href{mailto:j.pei2@vu.nl}{j.pei2@vu.nl}}
}

\graphicspath{{figs/}}
\begin{document}
\maketitle


\begin{abstract}        

Large language models (LLMs) can generate fluent dialogue, but prior works lack situational grounding, dynamic strategy control, and evaluation aligned with clinical standards in motivational interviewing (MI).
We introduce \textbf{StoryMI}, a multi-LLM agent framework for controllable MI dialogue generation, where questionnaire-based client profiles are expanded into situational stories that provide narrative context for the dialogue. Therapist and client agents generate MI-coded utterances guided by MI codes selected by the interaction agent, while an interaction agent dynamically coordinates exchanges to control MI strategies during a multi-turn conversation. We propose a two-level evaluation protocol: lexical metrics and MI-specific measures of macro-level counseling strategies, alongside LLM-as-judge and human expert assessments. We construct a dataset of 6K simulated MI dialogues grounded in 1K questionnaire-story pairs, covering 12 MI codes and 13 symptom domains, and benchmark six open- and closed-source LLMs. Our results show that situational grounding and macro-level control can improve MI adherence and clinical plausibility, demonstrating the effectiveness of a structured multi-agent workflow for psychotherapy dialogue generation. We provide code and data for reproducibility.\footnote{\url{https://github.com/Beren-sds/StoryMI}}

\end{abstract}


\begin{figure}[t]
    \centering
    \includegraphics[width=\columnwidth]{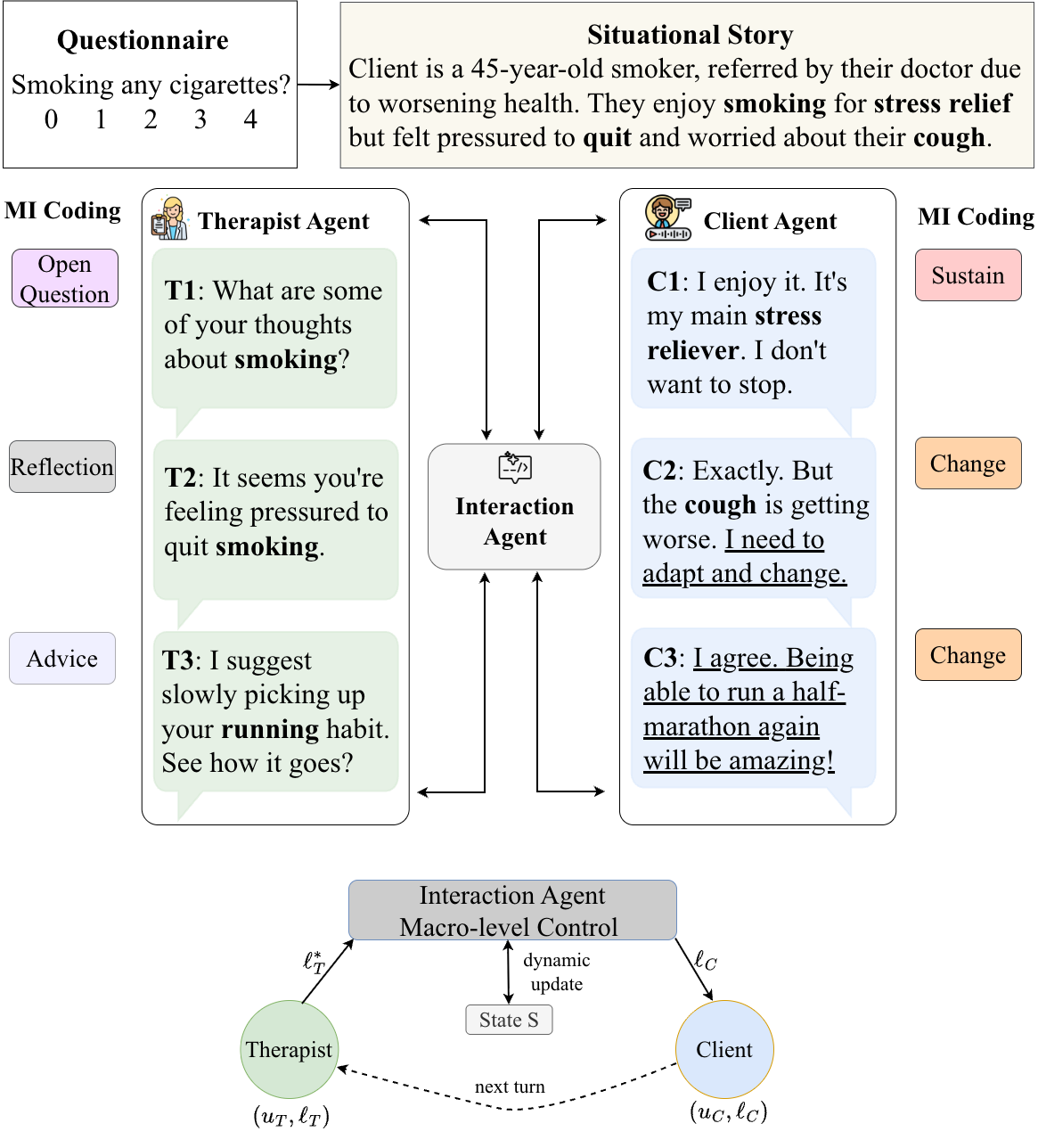}
    \caption{\OurModel{} workflow. 
    Situational story derived from questionnaire grounds MI-coded therapist--client dialogue.
    An interaction agent dynamically controls and coordinates MI-coded exchanges between therapist and client agents, with dynamic state updates.
    Bold and underlined spans indicate story overlap and intent to change.
    }
    \label{fig:story}
\end{figure}

\section{Introduction}\label{s:intro}

\Acf{MI} is a directive, client-centered counseling approach for eliciting clients' motivation for behavioral change~\citep{miller2012motivational}.
\Acf{BC}, which categorizes therapist and client behaviors~\citep{tavabi2021analysis}, mitigates client resistance and disengagement associated with confrontational or paternalistic interaction styles.
In addition, \textit{macro-level control} of a counseling strategy (a trajectory of behavior codes) is essential to guide the flow of the counseling~\citep{shah2022modeling}.
For example, in \Acf{MISC}~\citep{miller2000misc}, reflections should exceed questions by at least two to one; complex paraphrasing should dominate over simple echoing; and open questions should guide clients towards self-initiated change talk.

Unlike traditional methods~\citep{li2022knowledge, majumder-etal-2020-mime}, 
\acfp{LLM} enable psychotherapy dialogue agents through instruction following~\citep{ouyang2022training}, standard adherence~\citep{chiu2024computationalframeworkbehavioralassessment}, multi-party interaction simulation~\citep{park2023generative}.
However, three barriers remain:
\begin{enumerate*}
    \item \textit{Situational grounding}. 
DSM5AgentFlow~\citep{ozgun2025trustworthyaipsychotherapymultiagent} generates client profiles from questionnaires and simulates therapist--client agent conversations using an explainable diagnosis agent.
But Likert-scale ratings in questionnaires do not capture the client's situational context and personal intention, which are central to \acl{MI}.
    \item \textit{Dynamic controllability}. 
    Recent work includes automated \ac{BC}~\citep{wu2022annoMI,sun-etal-2024-eliciting} and \ac{MI}-guided dialogue generation~\citep{sun2025rethinking}, but MI codes remain static, script- or turn-level transformations rather than adaptive, situationally grounded responses. 
    Multi-LLM agent role-play frameworks~\citep{NEURIPS2023_CAMEL,qiu2024interactiveagents} produce fluent text but lack the structured control and fail to capture complex \acs{MI} interaction dynamics.
    \item \textit{Evaluation fidelity}. 
    Accessing the therapeutically meaningful dialogue requires more than surface-level coherence metrics, such as automatic lexical metrics~\citep{see-etal-2019-makes} and LLM-as-judge rubrics~\citep{zheng2023judgingllm-as-a-judge, liu-etal-2023-g}. None of the metrics captures macro-level control of counseling strategies (see \autoref{tab:related_work_comparison}), which is essential.
    
\end{enumerate*}

We introduce \OurModel{}, a multi-agent framework that grounds MI-coded client-therapist dialogue in situational stories derived from questionnaires, as shown in \autoref{fig:story}.
First, client profiles are generated from questionnaires and expanded into situational stories, providing rich contextual grounding for MI-coded dialogue.
Next, therapist and client agents alternately generate utterances along with their corresponding MI behavioral codes. 
An interaction management agent dynamically coordinates these exchanges, updating agent states to ensure adherence to MI strategies while maintaining naturalistic, situational conversation.

The contributions of this work are threefold:
\begin{itemize}
\item We design a multi-LLM agent workflow, which combines \textit{situational story contextualization} with dynamically \textit{controllable MI-coded interactions}, benchmarked across six representative LLMs.
\item We propose MI-specific metrics, together with existing automatic lexical metrics and LLM-as--judge rubrics, to quantify both micro and macro-level clinical counseling strategy.
\item We collect a dataset of 6,000 conversations grounded on 1,000 questionnaire-story pairs, covering 12 types of MI codes and 13 symptom domains for our evaluation and further study.
\end{itemize}

\section{Related Work}\label{s:background}

\subsection{MI-Adherent Dialogue Generation}

Prior work on MI dialogue generation focuses on utterance-level strategy modeling without grounding responses in client-specific situations. 
BiMISC~\citep{sun-etal-2024-eliciting} demonstrates that LLMs can predict MI coding categories and that explicit codes improve alignment at the utterance level, though the setup operates on isolated turns without multi-turn flows or client profiling. 
ScriptAlign~\citep{sun2025script} introduces script-strategy aligned generation using expert-crafted scripts paired with strategy labels, raising MI fidelity but limiting scalability due to handcrafted script dependence. 
RethinkMI~\citep{sun2025rethinking} proposes predicting MI strategies as intermediate reasoning before conditioning generation, improving controllability and explainability. DIIR~\citep{xie-etal-2024-shot-dialogue} produces a framework to apply MI conversation strategies in the form of inductive rules from expert demonstrations. VirturalMI~\citep{steenstra2024virtual} develops a virtual counselor to conduct MI counseling in the context of alcohol use. 

\OurModel{} addresses both gaps by replacing script dependence with standardized questionnaire profiling plus situational stories, enabling broad, clinically plausible coverage without manual script writing. Our controllable MI coding conditions each turn on behavioral strategy, while situational stories grounds responses in concrete client contexts.

\begin{table}[t!]
\centering
\caption{Comparison of most related work. 
The symbols \cmark, \pmark, \xmark~ indicate full support, partial support, and no support. 
CP = Client Profiling, TM = Therapist Modeling, MS = Multi-turn Strategy, DA = Dual Agent, MI = Motivational Interview, HE = Human Evaluation.
}
\label{tab:related_work_comparison}
\resizebox{\columnwidth}{!}{
\begin{tabular}{@{}lcccccc@{}}
\toprule
\textbf{Related Work} & CP & TM & MS & DA & MI & HE \\
\midrule
ChatPsych~\citep{chen2023llm} & \xmark & \xmark & \xmark & \pmark & \xmark & \cmark \\
PATIENT-$\Psi$~\citep{wang2024patient} & \pmark & \xmark & \pmark & \pmark & \xmark & \cmark \\
NoteChat~\citep{wang2023notechat} & \pmark & \xmark & \cmark & \pmark & \xmark & \xmark \\
ClientCAST~\citep{wang2024towards} & \xmark & \xmark & \cmark & \pmark & \xmark & \cmark \\
InteractAgent~\citep{qiu2024interactiveagents} & \xmark & \xmark & \cmark & \cmark & \xmark & \cmark \\
CPsyCoun~\citep{zhang2024cpsycoun} & \xmark & \xmark & \cmark & \xmark & \xmark & \cmark \\
MDD-5K~\citep{yin2025mdd5k} & \pmark & \xmark & \cmark & \cmark & \xmark & \xmark \\
DSM5AgentFlow~\citep{ozgun2025trustworthyaipsychotherapymultiagent} & \cmark & \xmark & \cmark & \cmark & \xmark & \cmark \\
AMIE~\citep{tu2024towards} & \pmark & \xmark & \cmark & \pmark & \xmark & \cmark \\
BiMISC~\citep{sun-etal-2024-eliciting} & \xmark & \cmark & \xmark & \xmark & \cmark & \cmark \\
ScriptAlign~\citep{sun2025script} & \xmark & \cmark & \cmark & \xmark & \cmark & \cmark \\
RethinkMI~\citep{sun2025rethinking} & \xmark & \cmark & \cmark & \xmark & \cmark & \cmark \\
\midrule
\textbf{\OurModel{} (Ours)} & \cmark & \cmark & \cmark & \cmark & \cmark & \cmark \\
\bottomrule
\end{tabular}
}
\end{table}

\subsection{Psycho-Conversation Simulation}

\paragraph{Single-Agent Systems.}
ChatPsych~\citep{chen2023llm} demonstrates tuning-free prompting for diagnostically competent chatbots but lacks modality-specific scaffolding and systematic multi-turn coding. PATIENT-$\Psi$~\citep{wang2024patient} develops CBT-style AI patients with cognitive distortions, yielding controllable personas yet remaining therapy-agnostic about MI mechanisms. NoteChat~\citep{wang2023notechat} conditions doctor-patient role-play on clinical notes for documentation quality rather than counseling strategy. These approaches model the therapist or client separately, lacking macro-level strategy control.

\paragraph{Multi-Agent Systems.} 
InteractAgent~\citep{qiu2024interactiveagents} role-plays counselor-client interactions and fine-tunes on synthetic data, while CPsyCoun~\citep{zhang2024cpsycoun} reconstructs multi-turn dialogues from clinical reports. 
MDD-5K~\citep{yin2025mdd5k} synthesizes psychiatrist-patient dialogues via neuro-symbolic diagnosis trees. 
DSM5AgentFlow~\citep{ozgun2025trustworthyaipsychotherapymultiagent} autonomously generates Likert-scale questionnaires and conducts simulations emphasizing diagnostic trustworthiness. 
These systems support multi-turn interaction but are neither controllable through explicit behavior coding nor situationally grounded in narrative client contexts.

\paragraph{Evaluation Fidelity.}
Almost all prior work relies on generic metrics (fluency, coherence) without alignment between automatic evaluation and clinical constructs. AMIE~\citep{tu2024towards} optimizes diagnostic medical dialogue through self-play but targets medical diagnosis rather than psychotherapy skill expression. ClientCAST~\citep{wang2024towards} benchmarks therapist performance through simulated client interactions but does not supply a generative recipe for MI-structured dialogues. StoryMI introduces specific metrics measuring macro-level strategies and systematically compares automatic, LLM-based, and human evaluation to quantify alignment on therapeutic dimensions.

\section{Workflow}\label{s:design}

\OurModel{} consists of three key components:
\begin{enumerate*}
    \item \textit{Questionnaire-Based Profiling} ($\S$\ref{s:questionnaire}) establishes psychologically consistent client representations with standard questionnaire assessment; 
    \item \textit{Situational Story Contextualization} ($\S$\ref{s:story}) bridges abstract symptom scores to narrative grounding, enabling the situational specificity that MI requires; and 
    \item \textit{Controllable MI Dialogue Simulation} ($\S$\ref{s:dialogue}) produces therapeutically coherent interactions through multi-agent coordination. 
    We simulate dialogue with three agents: client, therapist, and interaction manager. The interaction manager takes a special role of coordinating turns, selecting MI strategies, and enforcing therapeutic constraints via shared dialogue states.
\end{enumerate*}

\subsection{Questionnaire-Based Profiling}\label{s:questionnaire}
This module constructs structured client profiles from standardized clinical instruments to ensure consistent and controllable symptom presentation across dialogues.

We model client heterogeneity using the DSM-5 questionnaire~\citep{first2024dsm5,narrow2013dsm}, which contains 23 items spanning 13 symptom domains (e.g., depression, anger, sleep problems). 
For each item in the questionnaire, we use LLM generate (1) a severity score $s_i \in \{0,\ldots,4\}$ (from not at all to nearly every day), and (2) a brief first-person rationale describing the client's subjective experience.
The role-based system prompts and constrained JSON schemas are provided in \autoref{app:prompts}. The resulting structured profile $\mathcal{P}$ serves as an explicit, interpretable control signal that grounds downstream dialogue generation in stable psychological characteristics while supporting explainability and reproducibility~\citep{kim2025large}. A heterogeneity analysis of 1,000 client profiles is in \autoref{app:stats}.

\subsection{Situational Story Contextualization}\label{s:story}

This component transforms abstract questionnaire responses into situationally grounded narratives that enable naturalistic therapeutic exchange.

Raw questionnaire responses lack the contextual richness needed for naturalistic dialogue~\citep{gao-etal-2023-peacok,qiu2024interactiveagents}. For instance, a depression score of ``3'' indicates frequency but not circumstances, triggers, or personal meaning. We address this gap by generating a situational story $\mathcal{N}$ of approximately 200 words that transforms the client profile $\mathcal{P}$ into concrete life experiences. This narrative grounding enables persona consistency across multi-turn interactions, provides emotional texture for authentic therapeutic exchange, and supports scalable generation of psychologically plausible profiles without manual curation~\citep{kim-etal-2023-soda}. The story generation employs conditional prompting to select a primary symptom and construct a coherent first-person narrative around a specific scene, using constrained generation for symptom-behavior alignment.

\subsection{Controllable MI Dialogue Simulation}\label{s:dialogue}

\begin{algorithm}[t]
\footnotesize
\caption{MI-Coded Dialogue Generation}
\label{a:alg}
\begin{algorithmic}[1]
\Require Large language model $\mathcal{M}$, turn $ \in [T_{\min}, T_{\max}]$, context window $k$
\Statex \textbf{Initialize Dialogue State $\mathcal{S}$:}
\State $\mathcal{S}.\mathcal{H} \gets [\,]$ \Comment{Dialogue history: list of $(u, \ell)$ pairs}
\State $\mathcal{S}.\mathcal{C} \gets [\,]$ \Comment{MI code trajectory}
\State $\mathcal{S}.t \gets 0$ \Comment{Turn counter}
\State $\mathcal{S}.\psi \gets \text{False}$ \Comment{Completion flag}
\State $\mathcal{S}.\text{context} \gets (\mathcal{H}, \mathcal{C}, t, \mathcal{\psi})$
\Statex \textbf{Define Nodes $\mathcal{N}$:}
\State $\mathcal{N}_C$: ClientNode produces $(u_C, \ell_C)$ given $(\mathcal{S}, \mathcal{P}, \mathcal{N})$
\State $\mathcal{N}_T$: TherapistNode produces $(u_T, \ell_T)$ given $(\mathcal{S}, \ell_T^*)$
\State $\mathcal{N}_I$: InteractionNode selects $\ell_T^*$, updates $\mathcal{S}$, checks termination $\mathcal{S}.\psi$
\Statex \textbf{Questionnaire-Based Profiling \& Situational Story Contextualization}
\State $\mathcal{P} \gets \mathcal{N}_C.\text{FillQuestionnaire}(\mathcal{M})$ \Comment{Client profile}
\State $\mathcal{N} \gets \mathcal{N}_C.\text{GenerateStory}(\mathcal{P}, \mathcal{M})$ \Comment{Situational story}
\Statex \textbf{Controllable MI Dialogue Simulation}
\State $(u_0, \ell_0) \gets \mathcal{N}_T.\text{Greet}(\mathcal{M})$ \Comment{Initial greeting}
\State $\mathcal{S} \gets \mathcal{N}_I.\text{Update}(\mathcal{S}, u_0, \ell_0)$

\While{$\mathcal{S}.\psi \neq \text{True}$ \textbf{and} $\mathcal{S}.t < T_{\max}$}
    \State $(u_C, \ell_C) \gets \mathcal{N}_C.\text{Generate}(\mathcal{S}, \mathcal{M})$
    \State $\mathcal{S} \gets \mathcal{N}_I.\text{Update}(\mathcal{S}, u_C, \ell_C)$
    \State $\ell_T^* \gets \mathcal{N}_I.\text{SelectStrategy}(\mathcal{S}.\mathcal{C}[-k:], \ell_C, \mathcal{M})$ \Comment{Macro-level control}
    \State $(u_T, \ell_T) \gets \mathcal{N}_T.\text{Generate}(\mathcal{S}, \ell_T^*, \mathcal{M})$
    \State $\mathcal{S} \gets \mathcal{N}_I.\text{SynchronizeState}(\mathcal{S}, u_T, \ell_T)$
    \State $\mathcal{S}.t \gets \mathcal{S}.t + 1$ \Comment{Increment turn counter}
    \If{$\mathcal{S}.t \geq T_{\min}$}
        \State $\mathcal{S}.\psi \gets \mathcal{N}_I.\text{CheckTermination}(\mathcal{S}.\mathcal{H}, \mathcal{M})$
    \EndIf
\EndWhile
\State \Return $\mathcal{S}.\mathcal{H}$
\end{algorithmic}
\end{algorithm}

In this component, the interaction manager coordinates the client-therapist interactions with macro-level control and dynamic state updates. 
Algorithm~\ref{a:alg} formalizes controllable MI-coded dialogue generation using a multi-agent workflow. It first initializes the dialogue state and generates client profiles from questionnaires, which are then expanded into situational stories providing narrative grounding. Therapist and client agents alternately produce utterances paired with MI behavioral codes, while the interaction management agent dynamically selects strategies, updates states, and monitors termination. This loop continues until the dialogue reaches the maximum turn limit or meets the completion criteria, resulting in a fully simulated, MI-consistent conversation.

\subsubsection{MI Coding Scheme}\label{s:micoding}

We follow MISC/MITI~\citep{miller2003manualMISC,moyers2016motivationalMITI} scheme.
Therapist behaviors partition into three categories: 
\begin{enumerate*}
    \item \textit{Reflection} (simple vs.\ complex) captures empathic echoing or summarization, 
    \item \textit{Question} (open vs.\ closed) encompasses open-ended exploration and closed inquiries, and
    \item \textit{Input} includes information-giving, advice, affirmations, or goal-setting initiated by therapist.
\end{enumerate*}

Client utterances receive tri-partite motivational classification: 
\begin{enumerate*}
    \item \textit{Change} expresses desire, ability, or commitment toward behavioral change, 
    \item \textit{Sustain} articulates resistance or preference for status quo, and
    \item \textit{Neutral} constitutes non-motivational content discourse.
\end{enumerate*}

\subsubsection{MI-Conditioned Generation}

Each agent produces a dual output $(u, \ell)$ comprising both utterance and MI code. Generation proceeds in two stages. First, the Interaction Agent classifies the client's motivational orientation and selects an appropriate therapist strategy $\ell_T^* \in \mathcal{L}_T$ according to MI strategies. For instance, change talk typically receives reflective responses that reinforce motivation, while sustain talk calls for open questions exploring ambivalence. Second, the selected code conditions the therapist's response generation. This decoupling enables macro-level therapeutic control while preserving generation flexibility. The context is limited to the five most recent turns~\citep{yen-etal-2024-long}.

\subsubsection{Dynamic Interaction Management}

The interaction manager coordinates the client-therapist interactions with macro-level control and dynamic state updates. \autoref{fig:story} depicts the turn-by-turn coordination: the client agent produces an utterance with its MI code, the interaction agent classifies the client's MI state and selects an appropriate therapist strategy, and the therapist agent generates a response conditioned on this strategy.

This is implemented by three key functions in Algorithm~\ref{a:alg}:
\textit{SelectStrategy} examines recent client codes $\mathcal{C}[-k:]$ and selects a therapist strategy following MI strategies. 
\textit{SynchronizeState} updates the shared history $\mathcal{H}$ and code trajectory $\mathcal{C}$ after each turn, ensuring consistent context for generation. 
\textit{CheckTermination} monitors for natural closure indicators (gratitude, farewell) via LLM-based detection after $\tau \geq T_{\min}$ turns, with a hard bound at $T_{\max}$ to balance fidelity with computational feasibility~\citep{perez-rosas-etal-2016-building}.

\section{Evaluation Protocol}\label{s:eval}

We propose a two-level evaluation protocol: lexical metrics and MI-strategy metrics, alongside LLM-as-judge and human expert assessments.

\subsection{Lexical Metrics}

We employ four standard metrics capturing lexical fluency and diversity \citep{see-etal-2019-makes}.
\begin{enumerate*}
    \item \textit{Perplexity} measures fluency via language-model likelihood. Higher perplexity indicates less predictable, less templatic phrasing.
    \item \textit{Self-BLEU} quantifies intra-session redundancy by computing the average BLEU scores between each utterance and the rest of the dialogue. Lower values indicate less redundancy. 
    \item \textit{Distinct-2} measures the ratio of unique bi-grams to total bi-grams, capturing lexical diversity. 
    \item \textit{Entropy} quantifies token distribution uniformity across the dialogue.
\end{enumerate*}

\subsection{Strategy Metrics}
We introduce six behavioral metrics grounded in established MI quality indicators (See details in \autoref{app:metrics}). 
Here, the threshold $\delta$ follow the recommended macro-level strategies threshold defined in MISC coder's manual~\citep{miller2000misc}.
\begin{itemize}
    \item 
    \textit{Code Entropy}
    measures the diversity of therapeutic strategy usage via normalized Shannon entropy over the distribution of MI code categories. Higher values indicate balanced technique application across the defined strategies. 
    \item 
    \textit{Strategy Adherence}
    quantifies alignment with the golden MI strategy distribution, computed as the negative exponential of KL divergence from MISC best practices (50\% reflections, 25\% questions, 20\% input, 5\% other). 
    Scores near 1.0 indicate closer adherence to recommended practice patterns.
    \item \textit{Reflection Depth} quantifies how much semantic content a reflection adds beyond the client's original statement, computed as a weighted combination of sentence-level similarity (ensuring topical relevance) and token-level information gain (measuring novel therapeutic insight).
    \item \textit{Complex Reflection Ratio} ($\delta>0.5$) measures the proportion of reflections classified as paraphrase or summarization versus simple repetition, operationalizing reflection quality beyond surface-level echoing. 
    \item \textit{Open Question Ratio} ($\delta>0.7$) captures question type distribution, favoring exploratory open-ended inquiries over closed yes/no questions that limit client exploration. 
    \item \textit{Reflection-to-Question Ratio} ($\delta>2.0$) operationalizes the core MISC indicator that MI-adherent sessions should emphasize reflective responses over questioning, promoting client-driven exploration.
\end{itemize}

\subsection{LLM-as-Judge and Human Alignment}
We define six rubrics motivated by MI principles using 5-point Likert scales, to evaluate therapeutic dialogue quality for both LLM and human judges:
\textit{Coherence} (logical dialogue flow) \cite{dziri-etal-2019-evaluating}, \textit{Depth} (psychological insight and understanding), \textit{Progress} (therapeutic advancement towards resolution), \textit{Naturalness} (conversational authenticity), \textit{Empathy} (affective attunement to client state) \citep{sharma-etal-2020-computational, cabrera-lozoya-etal-2025-synthetic}, and \textit{Adherence} (adherence to MI strategies). 

Two expert annotators with clinical psychology backgrounds independently rated 120 dialogues (20 per model) using identical rubrics as the LLM-as-a-Judge evaluation. The stratified sampling ensured balanced representation across models. Inter-annotator reliability reached moderate agreement using quadratic-weighted Cohen's $\kappa$~\cite{artstein-poesio-2008-survey} ($\kappa = 0.51$ across all dimensions and models, with the highest reliability observed for Adherence as $\kappa = 0.79$).

\section{Experimental Setup}\label{s:experiments}
\subsection{Research Question}
We seek to answer three research questions in our evaluation:
\begin{enumerate*}[leftmargin=*, label=(\textbf{RQ\arabic*})]
    \item Can we simulate client-therapist conversations adhering to MI strategies with multiple LLM-based agents?
    \item How does situational story and MI codes influence the mental health conversation quality?
    \item How well does LLM-as-a-Judge align with human annotators with respect to metric and model evaluation levels? 
\end{enumerate*}

\begin{table*}[t]
\centering
\caption{Overall performance across two level metrics.
\textbf{Bold} and \underline{underline} indicate the best and second-best results.
\cmark~and \xmark~indicate whether a metric meets the recommended threshold.
$\uparrow$ higher is better; $\downarrow$ lower is better.}
\label{tab:automatic}
\resizebox{\textwidth}{!}{
\begin{tabular}{l >{\raggedleft\arraybackslash}p{1cm} >{\raggedleft\arraybackslash}p{1.2cm} >{\raggedleft\arraybackslash}p{1cm} >{\raggedleft\arraybackslash}p{1.2cm} >{\raggedleft\arraybackslash}p{1.2cm} >{\raggedleft\arraybackslash}p{1cm} >{\raggedleft\arraybackslash}p{1.0cm} >{\raggedleft\arraybackslash}p{0.1cm}}
\toprule
\textbf{Metric} & \multicolumn{1}{c}{\textbf{GPT-5-Nano}} & \multicolumn{1}{c}{\textbf{LLaMA 3.1-8B}} & \multicolumn{1}{c}{\textbf{Phi-4-14B}} & \multicolumn{1}{c}{\textbf{OpenChat-7B}} & \multicolumn{1}{c}{\textbf{Gemma-7B}} & \multicolumn{1}{c}{\textbf{Qwen 2.5-7B}} & \multicolumn{1}{c}{\textbf{Overall}}\\
\midrule
\multicolumn{8}{@{}l}{\textit{Lexical Metrics}} \\
Entropy (\%) $\uparrow$ & \textbf{90.7} & 87.5 & 88.6 & 89.4 & \underline{89.3} & 89.1 & 89.1 \\
Distinct-2 (\%) $\uparrow$ & \textbf{84.3} & 75.7 & \underline{76.8} & 69.8 & 64.0 & 73.0 & 73.9 \\
Perplexity $\uparrow$ & \underline{12.2} & 10.9 & \textbf{17.1} & 7.7 & 7.3 & 10.0 & 10.9 \\
Self-BLEU (\%) $\downarrow$ & \textbf{16.7} & 27.0 & \underline{25.5} & 39.3 & 49.3 & 34.2 & 32.0 \\
\midrule
\multicolumn{8}{@{}l}{\textit{Strategy Metrics}} \\
Code Entropy (\%) $\uparrow$ & \underline{80.0} & 83.2 & \textbf{87.9} & 86.6 & 79.7 & 72.3 & 81.6 \\
Strategy Adherence (\%) $\uparrow$ & \underline{80.9} & 53.0 & 53.0 & \textbf{81.3} & 80.1 & 81.0 & 71.6 \\
Reflection Depth (\%) $\uparrow$ & \underline{72.5} & 45.7 & 42.5 & 67.0 & 72.1 & \textbf{75.9} & 62.6 \\
Complex Reflection Ratio ($\delta$>50\%) & \ymark 98.3 & \ymark 62.2 & \ymark 58.0 & \ymark 82.6 & \ymark 97.4 & \ymark 99.0 & \ymark 83.0 \\
Open Question Ratio ($\delta$>70\%) & \nmark\phantom{0}6.9 & \ymark 91.8 & \ymark\textbf{95.2} & \nmark 11.7 & \nmark 29.0 & \nmark 19.8 & \nmark 42.4 \\
Reflection/Question Ratio ($\delta$>2.0) & \ymark\phantom{0}3.6 & \nmark\phantom{0}0.1 & \nmark\phantom{0}0.5 & \nmark\phantom{0}1.9 & \ymark\phantom{0}4.4 & \ymark\phantom{0}\textbf{6.3} & \ymark\phantom{0}2.8 \\
\bottomrule
\end{tabular}
}
\end{table*}

\subsection{Dataset and LLMs}

We construct a synthetic dataset of 6,000 multi-turn dialogues (113K+ utterances) grounded in 1,000 questionnaire--story pairs.
Each questionnaire covers 23 DSM-5 items across 13 symptom domains. Each story averages around 200 words. 
Average dialogue lengths range from 13.3 to 25.6 turns, with utterance lengths from 27 words to 133 words. 
Detailed statistics are provided in~\autoref{tab:dialogue-stats}.
We use seven LLMs, including GPT-5-Nano\footnote{\url{https://platform.openai.com/docs/models/gpt-5-nano}} and five open-source \acp{LLM} provided by Ollama,\footnote{\url{https://ollama.com/}} i.e., LLaMA 3.1-8B, Qwen 2.5-7B, Gemma-7B, OpenChat-7B, and Phi-4-14B (hereafter, GPT, LLaMA, Phi, OpenChat, Gemma, Qwen), and GLM-5 (short for GLM) as an independent evaluator to rule out evaluator bias.

\subsection{Implementation details}

We use LangGraph\footnote{\url{https://www.langchain.com/langgraph}} to develop StoryMI (\autoref{app:time_complexity}).
All 1,000 questionnaires and background stories were generated once using the same base model LLaMA with fixed decoding parameters and reused for all six generation models.
In the ablation studies (without story, without MI code, both removed), all other decoding and inference settings for GPT stay unchanged compared with the full condition.
We set the temperature to $0.7$ and top-p to $0.9$ for open-source \acp{LLM}. LLaMA served as the reference model for generating all the questionnaire profiles and situational stories. We use GPT as the closed-source \ac{LLM}, which also serves as the LLM-as-judge evaluator.
We run open-source \acp{LLM} on the region-based national supercomputer cluster using a single NVIDIA A100 GPU (40GB).

\section{Results}

\subsection{Overall Performance (RQ1)}
\subsubsection{Lexical and Strategy Evaluation} 
\autoref{tab:automatic} reports overall performance across lexical diversity and MI-specific strategy metrics.

First, StoryMI enables surface-level language quality across diverse LLMs.
GPT attains the highest Entropy (90.7\%) and Distinct-2 (84.3\%), alongside the lowest Self-BLEU (16.7\%), indicating both high diversity and low repetition. 
Open-source models also demonstrate competitive fluency (e.g., Phi achieves 88.6\% Entropy and 17.1 Perplexity), indicating that they enables fluent and diverse dialogue generation across model families.

Second, StoryMI consistently enables macro-level MI strategy control.
Strategy Adherence is high (>80\%) for most models (GPT, OpenChat, Gemma, and Qwen), indicating close alignment with the target MI strategy distribution.
The Complex Reflection Ratio further demonstrates adherence to MI strategies, with all models exceeding the recommended threshold of 50\%.
These results directly validate StoryMI's interaction agent as an effective mechanism for enforcing MI-adherent macro-strategies beyond turn-level prompting.

Third, lexical metrics alone fail to explain therapeutic quality differences.
For instance, Phi shows the highest Perplexity (17.1) and Code Entropy (87.9\%) but low Strategy Adherence (53.0\%) and shallow Reflection Depth (42.5\%). 
In contrast, Qwen achieves the highest Reflection Depth (75.9\%) and Complex Reflection Ratio (99.0\%) despite moderate lexical scores. 
This divergence demonstrates that MI-specific metrics reveal clinically meaningful distinctions that surface-level metrics cannot capture, underscoring the necessity of StoryMI's evaluation framework.

\subsubsection{LLM-as-Judge and Human Evaluation}

\begin{figure}[h!]
    \centering
    \begin{subfigure}{0.485\columnwidth}
        \centering
        \includegraphics[width=\linewidth]{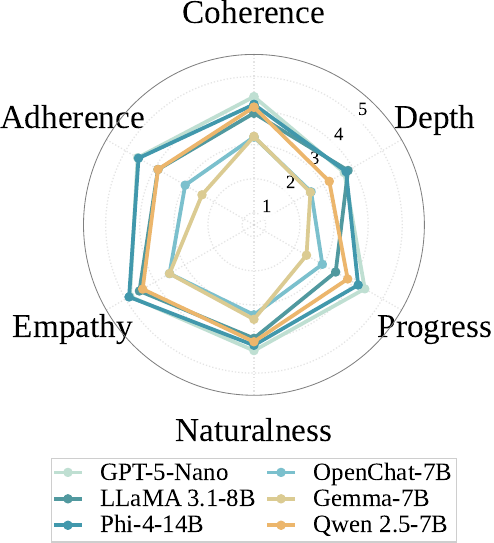}
        \caption{LLM-as-a-Judge}
        \label{fig:radar(a)}
    \end{subfigure}
    \hfill
    \begin{subfigure}{0.485\columnwidth}
        \centering
        \includegraphics[width=\linewidth]{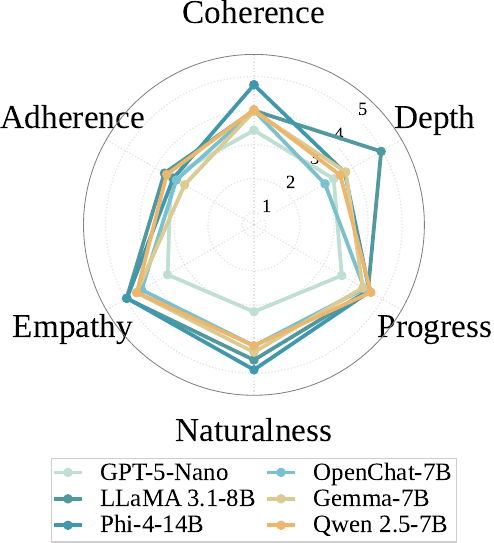}
        \caption{Human evaluation}
        \label{fig:radar(b)}
    \end{subfigure}
    \caption{Therapeutic dialogue quality evaluation in Likert scale (1 to 5 points) across six dimensions.}
    \label{fig:radar_all}
\end{figure}

\autoref{fig:radar_all} presents rubric-based evaluation from both LLM judges and human experts.

First, Depth shows the strongest cross-evaluator consistency.
Both LLM judges and human experts assign high Depth scores to dialogues with substantive psychological exploration, with LLaMA-generated dialogues receiving the highest ratings from both evaluator types (LLM: 3.84; Human: 4.96). This agreement suggests that Depth captures observable dialogue properties that both automated and human evaluation can reliably assess.

Second, Naturalness and Adherence show the largest cross-evaluator divergence.
LLM judges produce uniformly high scores across models (e.g., GPT Adherence: 4.59), while human experts assign more varied ratings with notably lower scores on certain dimensions (e.g., GPT Naturalness: 3.20). This pattern indicates that Naturalness and Adherence require assessment of subtle qualities, such as conversational authenticity and technique fidelity, that LLM judges have difficulty capturing.

Third, the Open Question Ratio correlates with higher human Empathy ratings.
Models exceeding the 70\% Open Question Ratio threshold (LLaMA: 91.8\%; Phi: 95.2\%) receive the highest human Empathy scores (4.97 and 4.95, respectively). Models with lower open question ratios (GPT: 6.9\%; OpenChat: 11.7\%) receive correspondingly lower Empathy ratings. This correlation supports the MI principle that open questions embody an accepting therapeutic stance, and demonstrates that \OurModel{}'s controllable generation enables systematic investigation of technique-outcome relationships.

\subsection{Ablation Study (RQ2)}

We conduct an ablation study of the key components of \OurModel{}, as shown in \autoref{tab:ablation}.

First, MI coding drives strategy adherence. 
Removing MI codes drops Strategy Adherence by 16.7 points (80.9\% to 64.2\%) while lexical metrics remain highly stable. This confirms that interaction-level coding enables macro-level strategy control.
Second, situational stories improve behavioral diversity.
Removing stories reduces Code Entropy by 5.0 points while maintaining Strategy Adherence, indicating that narrative grounding enriches therapeutic exchanges between client and therapist.
Third, combined removal produces the largest degradation.
Strategy Adherence drops from 80.9\% to 61.4\%, demonstrating that both components contribute complementary benefits (qualitative examples in \autoref{app:qualitative}).

\begin{table}[t]
\centering
\scriptsize
\setlength{\tabcolsep}{2pt}
\caption{Ablation study on GPT-5-Nano. \textbf{Bold} = largest degradation for each metric.}
\label{tab:ablation}
\resizebox{\columnwidth}{!}{
\begin{tabular}{@{}lcccc@{}}
\toprule
\textbf{Metric} & \textbf{Full} & \textbf{w/o Story} & \textbf{w/o MI} & \textbf{w/o Both} \\
\midrule
\multicolumn{5}{@{}l}{\textit{Lexical Metrics}} \\
Entropy (\%) $\uparrow$ & 90.7 & 90.9 & \textbf{90.4} & 90.8 \\
Distinct-2 (\%) $\uparrow$ & 84.3 & 84.7 & 84.5 & 85.5 \\
Perplexity $\uparrow$ & 12.2 & 12.8 & \textbf{12.1} & 12.8 \\
Self-BLEU (\%) $\downarrow$ & 16.7 & 15.4 & 17.6 & \textbf{15.0} \\
\midrule
\multicolumn{5}{@{}l}{\textit{Strategy Metrics}} \\
Code Entropy (\%) $\uparrow$ & 80.0 & 75.0 & \textbf{69.9} & 70.1 \\
Strategy Adherence (\%) $\uparrow$ & 80.9 & 78.8 & 64.2 & \textbf{61.4} \\
Reflection Depth (\%) $\uparrow$ & 72.5 & 73.8 & 68.5 & \textbf{63.4} \\
Complex Reflection Ratio ($\delta >50\%$) & 98.3 & 98.6 & 90.5 & \textbf{82.3} \\
Open Question Ratio ($\delta >70\%$) & \phantom{0}6.9 & \textbf{\phantom{0}4.7} & \phantom{0}6.6 & 23.2 \\
Reflection/Question Ratio ($\delta >2.0$) & \phantom{0}3.6 & \phantom{0}4.8 & \phantom{0}5.3 & \phantom{0}3.9 \\
\bottomrule
\end{tabular}
}
\end{table}

\subsection{Human Alignment Analysis (RQ3)}
\label{sec:human_alignment_analysis}

\begin{figure}[htb!]
    \centering 
    \begin{subfigure}[b]{\columnwidth}
        \centering
        \includegraphics[width=1\columnwidth]{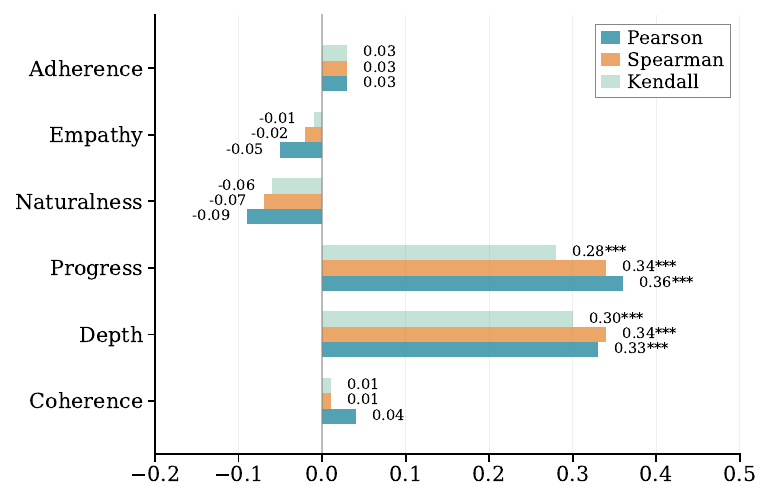}
        \caption{Dimension-level Correlation}
        \label{fig:correlation(a)}
    \end{subfigure}
    \begin{subfigure}[b]{\columnwidth}
        \centering
        \includegraphics[width=1\columnwidth]{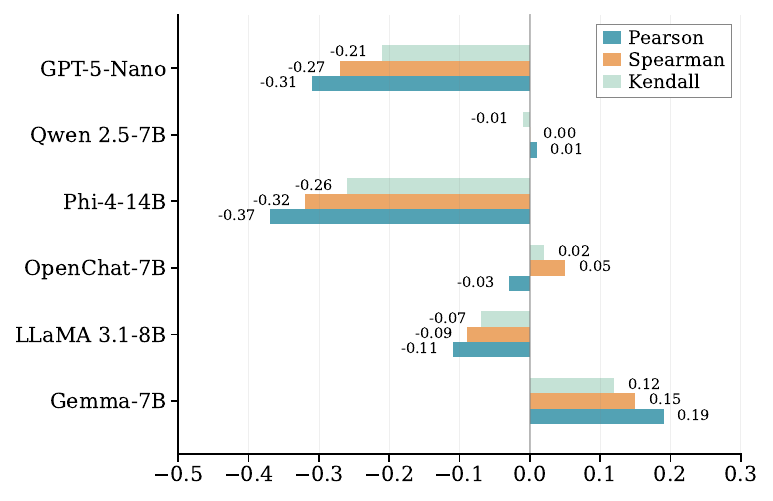}
        \caption{Model-level Correlation}
        \label{fig:correlation(b)}
    \end{subfigure}
    \caption{Correlation between LLM-based and human evaluations.
    Significance levels: * $p < 0.05$, ** $p < 0.01$, *** $p < 0.001$.}
    \label{fig:correlation_all}
\end{figure}

\autoref{fig:correlation_all} presents correlation analysis between LLM-based and human evaluations.

First, Depth and Progress show significant positive correlations.
Depth ($r$=0.33) and Progress ($r$=0.36) indicate that LLM judges can reliably identify dialogues that explore issues meaningfully and advance logically. These dimensions assess observable dialogue properties: whether conversations move beyond surface remarks and progress without repetition.

Second, Coherence, Naturalness, Empathy, and Adherence show near-zero correlations.
Despite high human inter-rater reliability on Adherence ($\kappa$=0.79), in general LLM-human correlations remain weak ($|r|$<0.10). These dimensions require assessment of subtler qualities: conversational flow, authenticity, affective attunement, and therapeutic technique fidelity.

Third, no model achieves strong cross-evaluator consistency.
All models show weak correlations ($|r|$<0.38), indicating that LLM-human agreement on relative dimension rankings is limited regardless of overall dialogue quality. 
Overall, LLM-as-Judge can efficiently screen for dialogue development quality at scale, while human experts remain necessary for validating therapeutic fidelity and conversational authenticity.

\subsection{Self- vs Cross-Evaluator Bias Analysis}
\label{subsec:bias}
\paragraph{Bias Verfication.}
To rule out evaluator bias, we re-evaluated all GPT-generated dialogues with GLM using the same rubric and conducted a paired $t$-test against the self-evaluation scores, as shown in \autoref{tab:cross_eval_self}.

We find that GPT rates its own dialogues \emph{lower} than GLM on 5 of 6 dimensions ($p < 0.001$), indicating no inflated scoring tendency. Depth is the sole exception (+0.19), where GPT gives itself a slightly higher score.

\paragraph{Cross-Model Ranking Consistency.}
\autoref{tab:cross_eval_rank} compares the two evaluators' results across six models, including their ranking positions.
The top-4 and bottom-2 model clusters are consistent; notably, GLM ranks Phi first and GPT second, confirming that the original ranking holds under an independent evaluator.
Furthermore, the top-4 and bottom-2 model ranking clusters remain consistent across two evaluators.
\begin{table}[htb!]
\centering
\small
\caption{Comparison of self- and cross-evaluator results across different rubrics, including a paired $t$-test of their differences. Diff.\ = GPT - GLM).}
\label{tab:cross_eval_self}
\resizebox{\columnwidth}{!}{
\begin{tabular}{@{}lccccc@{}}
\toprule
 & \multicolumn{2}{c}{\textbf{LLM Evaluators}}  & &   \multicolumn{2}{c}{\textbf{Paired $t$-test}} \\
& \textbf{GPT-5-Nano} & \textbf{GLM-5} & \textbf{Diff.} & $t$ & $p$ \\
\midrule
Coherence & 4.48 & 4.93 & $-$0.45 & $-$8.35 & {<}.001 \\
Depth & 3.74 & 3.55 & \textbf{+0.19} & 3.27 & .001 \\
Progress & 4.46 & 4.94 & $-$0.48 & $-$9.20 & {<}.001 \\
Naturalness & 4.39 & 4.81 & $-$0.42 & $-$6.75 & {<}.001 \\
Empathy & 4.90 & 4.98 & $-$0.08 & $-$2.36 & .020 \\
MI Alignment & 4.71 & 4.94 & $-$0.23 & $-$4.35 & {<}.001 \\
\midrule
\textbf{Overall} & \textbf{4.45} & \textbf{4.69} & \textbf{$-$0.24} & $-$6.03 & {<}.001 \\
\bottomrule
\end{tabular}
}
\end{table}

\begin{table}[htb!]
\centering
\small
\caption{Comparison of the two evaluators’ alignment with human annotations, measured using Pearson’s $r$, Spearman’s $\rho$, and Kendall’s $\tau$ correlation coefficients. Diff.\ = GPT-GLM).
}
\label{tab:cross_eval_rank}
\resizebox{\columnwidth}{!}{
\begin{tabular}{@{}lccccc@{}}
\toprule
 & \multicolumn{2}{c}{\textbf{LLM Evaluators}}  & &   \multicolumn{2}{c}{\textbf{Rank Position}} \\
\textbf{Model} & \textbf{GPT-5-Nano} & \textbf{GLM-5} & \textbf{Diff.} & \textbf{GPT-5-Nano} & \textbf{GLM-5} \\
\midrule
GPT-5-Nano & 4.45 & 4.69 & $-$0.24 & 1 & \textbf{2} \\
Phi-4-14B & 4.31 & \textbf{4.78} & $-$0.47 & 2 & \textbf{1} \\
Qwen-2.5-7B & 3.94 & 4.17 & $-$0.23 & 3 & 4 \\
LLaMA-3.1-8B & 3.93 & 4.49 & $-$0.57 & 4 & 3 \\
OpenChat-7B & 3.20 & 2.83 & +0.37 & 5 & 5 \\
Gemma-7B & 2.82 & 2.59 & +0.23 & 6 & 6 \\
\bottomrule
\end{tabular}
}
\end{table}

\paragraph{Human Alignment Consistency.}\label{sec:human_alignment}

\autoref{tab:cross_eval_corr} compares the alignment of two evaluators’ judgments with human ratings.
GLM achieves higher human correlation on 4 of 6 dimensions, averaging across three correlation metrics. Progress is particularly notable ($r = 0.54$ vs.\ $0.43$). On Naturalness, GPT's correlation is near zero ($r = 0.01$), while GLM reaches significance ($\rho = 0.22$, $p < .05$), consistent with our finding that Naturalness shows the largest human--LLM gap (\S\ref{sec:human_alignment_analysis}). Depth is the only dimension where GPT correlates more strongly, possibly reflecting familiarity with its own generation patterns.

\begin{table}[h!]
\centering
\small
\caption{
Comparison of the two evaluators' judgment alignment with human annotations, measured by Pearson’s $r$, Spearman’s $\rho$, and Kendall’s $\tau$ correlation coefficients.
Significance: $^*p<.05$, $^\dagger p<.01$, $^\ddagger p<.001$.}
\label{tab:cross_eval_corr}
\setlength{\tabcolsep}{10pt}
\resizebox{\columnwidth}{!}{
\begin{tabular}{@{}lccccccl@{}}
\toprule
 & \multicolumn{3}{c}{\textbf{GLM-5}} & \multicolumn{3}{c}{\textbf{GPT-5-Nano}}\\
\textbf{Dimension} & \textbf{$r$} & \textbf{$\rho$} & \textbf{$\tau$} & \textbf{$r$} & \textbf{$\rho$} & \textbf{$\tau$} \\
\midrule
Coherence & .128 & .157 & .134 & .098 & .100 & .087 \\
Depth & .313$^\dagger$ & .314$^\dagger$ & .267$^\dagger$ & .368$^\ddagger$ & .384$^\ddagger$ & .336$^\ddagger$ \\
Progress & .538$^\ddagger$ & .504$^\ddagger$ & .408$^\ddagger$ & .432$^\ddagger$ & .417$^\ddagger$ & .351$^\ddagger$ \\
Naturalness & .166 & .219$^*$ & .182$^*$ & .013 & .039 & .036 \\
Empathy & .034 & .049 & .047 & .012 & .082 & .075 \\
MI Alignment & .169 & .172 & .148 & .111 & .156 & .137 \\
\bottomrule
\end{tabular}
}
\end{table}

\section{Discussion and Implication}

\paragraph{Complementary Evaluation Paradigms.}
Our correlation analysis shows that LLM judges and human experts capture different aspects of therapeutic dialogue quality. LLM evaluators reliably assess observable properties such as topical exploration and logical progression, while human experts prove necessary for subtler qualities including conversational authenticity and affective attunement. This divergence suggests a practical division of labor: LLM-based methods can efficiently screen dialogue development at scale~\citep{zheng2023judgingllm-as-a-judge,liu-etal-2023-g}, while human judgment remains necessary for validating therapeutic fidelity~\citep{basar-etal-2025-well}. We advocate for hybrid protocols that use LLM efficiency for initial filtering while reserving expert assessment for dimensions where automated evaluation falls short, such as empathy and naturalness as observed in our study.

\paragraph{Open Questions and Therapeutic Stance.}
Our findings show a consistent relationship between open question usage and human-perceived empathy: models with higher open question ratios receive higher empathy ratings from human experts. This aligns with MI theory, which positions therapeutic empathy as emerging from an accepting, exploratory stance~\citep{miller2012motivational}. Open questions create space for client elaboration and signal genuine interest in the client's perspective, which human raters appear to recognize more readily than LLM judges. The controllable generation in \OurModel{} allows investigation of how MI strategies relate to human perceived quality, further extending prior work on strategy-aligned generation~\citep{sun2025script}.

\paragraph{Multiple Paths to Therapeutic Modeling.}
Our evaluation shows that models achieve therapeutic plausibility through distinct strategic profiles: some excel through exploratory questioning while others emphasize reflective techniques. Both paths can lead to favorable human ratings, suggesting that therapeutic quality is not tied to a single behavioral pattern. Specifically, strategy metrics ensure technique fidelity to MI principles~\citep{moyers2016motivationalMITI}, while human evaluation validates perceived therapeutic effectiveness. This complementarity supports multi-dimensional assessment that considers both behavioral adherence and subjective quality~\citep{see-etal-2019-makes}.

\paragraph{Client Profiling and Persona Diversity.}
Relying on standardized questionnaire instruments inherently constrains the persona space: the resulting client profiles are shaped by the fixed item set and Likert-scale format of the DSM-5 cross-cutting symptom measures, which may not fully capture idiosyncratic personal narratives, cultural expressions of distress, or contextual factors beyond symptom frequency. This profiling choice propagates through the pipeline, influencing both the diversity of situational stories and the range of therapeutic dynamics represented in the generated dataset. Future work could explore complementary profiling strategies, such as open-ended intake interviews, to broaden persona coverage.

\paragraph{Implications.}
MI-style therapeutic quality can be shaped by controlling dialogue strategies rather than relying on model scale alone. Different strategy profiles achieve comparable human-rated quality, suggesting MI systems should adapt to context rather than follow rigid recipes. Evaluation frameworks should jointly report strategy adherence and human perception, since either alone misses key aspects of therapeutic stance.

\section{Conclusion}\label{s:conclusion}
This work shows that structured multi-agent coordination enables controllable and situationally grounded MI dialogue generation.
\OurModel{} integrates questionnaire-based client profiling, narrative contextualization, and macro-level MI behavioral coding to produce therapeutically coherent dialogues.
Our evaluation demonstrates that MI-specific metrics capture clinically meaningful distinctions beyond lexical measures alone. The workflow, dataset, and evaluation protocol together provide a foundation for future research, with open directions including extension to validation in real-world training and clinical settings.

\section*{Limitations}
We acknowledge several limitations. 
First, our framework focuses on motivational interviewing as a single therapeutic modality, whereas real clinical practice often integrates multiple approaches; extending controllable generation to multi-modality therapy would require additional coding schemes beyond MISC. 
Second, the dialogues are validated through expert annotation rather than interaction with actual clients or deployment in clinical training settings, so user studies with trainees or clinicians remain necessary to establish real-world applicability. 
Third, the questionnaire-based profiling reflects Western diagnostic norms embedded in DSM-5 and may require adaptation for cross-cultural contexts where MI is practiced with different populations. 
Finally, while we demonstrate LLM--human alignment patterns, establishing causal relationships between specific MI strategies and therapeutic outcomes would require longitudinal studies with actual client populations, which is beyond the scope of this work.

\bibliography{references}

\appendix

\section{MI Strategy Metric Definition}\label{app:metrics}

This appendix provides complete mathematical definitions, computational details, and illustrative examples for the proposed MI-specific automatic metrics introduced in $\S$\ref{s:eval}.

\subsection{MI Code Distribution Entropy}

MI Code Entropy measures the diversity of therapeutic strategies employed throughout a dialogue session. Higher entropy indicates balanced strategy usage across MI techniques; lower entropy suggests over-reliance on specific strategies.

\begin{sloppypar}
\paragraph{Definition.} Given a sequence of therapist MI codes $\mathbf{c} = (c_1, \ldots, c_n)$ where each $c_i \in \mathcal{C} = \{\texttt{reflection}, \allowbreak \texttt{question}, \allowbreak \texttt{input}, \allowbreak \texttt{other}\}$, we compute:
\end{sloppypar}

\begin{equation}
H(\mathbf{c}) = -\frac{1}{\log_2 |\mathcal{C}_{\text{obs}}|} \sum_{c \in \mathcal{C}_{\text{obs}}} p(c) \log_2 p(c)
\end{equation}

\begin{sloppypar}
where $p(c) = \text{count}(c)/n$ and $\mathcal{C}_{\text{obs}} \subseteq \mathcal{C}$ is the set of observed categories. We normalize by $\log_2 |\mathcal{C}_{\text{obs}}|$ to ensure $H \in [0, 1]$.
\end{sloppypar}

\begin{sloppypar}
\paragraph{Example.} Consider a 10-turn dialogue with the following counts: \texttt{reflection} (6), \texttt{question} (3), \texttt{therapist\_input} (1), and \texttt{other} (0).
\end{sloppypar}

\begin{sloppypar}
The observed set is $\mathcal{C}_{\text{obs}} = \{\texttt{reflection}, \allowbreak \texttt{question}, \allowbreak \texttt{input}\}$, so $|\mathcal{C}_{\text{obs}}| = 3$. The probabilities are $p(\texttt{refl}) = 0.6$, $p(\texttt{ques}) = 0.3$, and $p(\texttt{input}) = 0.1$.
\end{sloppypar}

The raw entropy calculation is:
\begin{align*}
H_{\text{raw}} &= -(0.6 \log_2 0.6 + 0.3 \log_2 0.3 \\
&\quad + 0.1 \log_2 0.1) \\
&\approx 1.295
\end{align*}

The final normalized score is $H = 1.295 / \log_2(3) = 0.817$.

\subsection{Strategy Adherence}

The Strategy Adherence ($\mathcal{S}_{adh}$) quantifies how closely the observed distribution of MI codes match best-practice recommendations.

\paragraph{Ideal Distribution.} Based on MISC guidelines we define:
\begin{align*}
P_{\text{ideal}} = \{&\texttt{reflection}: 0.50, \texttt{question}: 0.25, \nonumber \\
&\texttt{input}: 0.20, \texttt{other}: 0.05\}
\end{align*}

\begin{sloppypar}
\paragraph{Definition.} Given observed distribution $P_{\text{obs}}$, we compute:
\end{sloppypar}

\begin{equation}
\mathcal{S}_{adh} = \exp\left(-D_{\text{KL}}(P_{\text{obs}} \| P_{\text{ideal}})\right)
\end{equation}

\begin{sloppypar}
We apply additive smoothing ($\epsilon = 10^{-6}$) to handle zero counts.
\end{sloppypar}

\begin{sloppypar}
\paragraph{Example.} Using the previous counts, $P_{\text{obs}} = \{0.6, 0.3, 0.1, \epsilon\}$. The KL divergence is:
\end{sloppypar}

\begin{align*}
D_{\text{KL}} &\approx 0.6 \ln(0.6/0.5) + 0.3 \ln(0.3/0.25) \\
&\quad + 0.1 \ln(0.1/0.2) \\
&\approx 0.109 + 0.055 - 0.069 = 0.095
\end{align*}

The final score is $\mathcal{S}_{adh} = \exp(-0.095) = 0.909$.

\subsection{Reflection Depth}

\begin{sloppypar}
Reflection Depth ($\mathcal{R}_d$) computes a continuous score representing the semantic elaboration of a reflection beyond the client's original statement. It relies on two components: Semantic Similarity and Information Gain.
\end{sloppypar}

\paragraph{Semantic Similarity ($\mathcal{\text{Sim}}$).} We encode the reflection $r$ and client utterance $u$ into embeddings $\mathbf{e}_r, \mathbf{e}_u$ using the \texttt{all-MiniLM-L6-v2} model. The raw cosine similarity is normalized to $[0, 1]$:
\begin{equation}
\mathcal{\text{Sim}}(r, u) = \frac{1}{2} \left( \frac{\mathbf{e}_r \cdot \mathbf{e}_u}{\|\mathbf{e}_r\| \|\mathbf{e}_u\|} + 1 \right)
\end{equation}

\begin{sloppypar}
\paragraph{Information Gain ($\mathcal{\text{Info}}$).} We define $T_r$ and $T_u$ as the sets of lemmatized content tokens (excluding stopwords/punctuation) for $r$ and $u$. A token $t \in T_r$ is considered \textit{novel} if it is semantically distinct from all tokens in $T_u$:
\end{sloppypar}
\begin{equation}
\text{IsNovel}(t) = \mathbb{I}\left(\max_{t' \in T_u} \text{cossim}(t, t') < \theta\right)
\end{equation}
where $\text{cossim}(t, t')$ is the cosine similarity between token embeddings and the threshold $\theta = 0.8$. We compute:
\begin{equation}
\mathcal{\text{Info}}(r, u) = \frac{\sum_{t \in T_r} \text{IsNovel}(t)}{|T_r|}
\end{equation}

\paragraph{Definition.} The final score is a weighted sum:
\begin{equation}
\mathcal{R}_d = \frac{1}{|R|} \sum_{(r, u) \in R} (0.4 \cdot \mathcal{\text{Sim}}(r, u) + 0.6 \cdot \mathcal{\text{Info}}(r, u))
\end{equation}

\begin{sloppypar}
\paragraph{Rationale.} We weight Information Gain higher (0.6) to reward reflections that add therapeutic meaning (paraphrasing) rather than merely repeating content.
\end{sloppypar}

\begin{sloppypar}
\paragraph{Example.}
\textit{Client ($u$):} ``I feel so tired.'' ($T_u = \{\text{feel}, \text{tired}\}$)
\textit{Reflection ($r$):} ``It sounds like you are exhausted.'' ($T_r = \{\text{sound}, \text{exhausted}\}$)
\end{sloppypar}
\begin{itemize}
    \setlength\itemsep{0em}
    \item \textbf{$\mathcal{\text{Sim}}$:} Embeddings are close but not identical; assume normalized $\mathcal{\text{Sim}} \approx 0.70$.
    \item \textbf{$\mathcal{\text{Info}}$:} ``Exhausted'' is similar to ``tired'' (cossim $> 0.8$), so it is \textit{not} novel. ``Sound'' is novel. $\mathcal{\text{Info}} = 1/2 = 0.5$.
    \item \textbf{Score:} $0.4(0.7) + 0.6(0.5) = 0.28 + 0.30 = 0.58$.
\end{itemize}

\subsection{Complex Reflection Ratio}
\begin{sloppypar}
The Complex Reflection Ratio ($\mathcal{R}_c$) measures the proportion of reflections demonstrating therapeutic depth beyond simple repetition.
\end{sloppypar}

\begin{sloppypar}
\paragraph{Metrics Calculation.} For a reflection $r$ and client utterance $u$,
we compute:
\begin{enumerate*}
    \item \textbf{Similarity ($\mathcal{\text{Sim}}$)}: Cosine similarity of sentence embeddings (\texttt{all-MiniLM-L6-v2} model, mapping to 384 dimensional dense vector space).
    \item \textbf{Info Gain ($\mathcal{\text{Info}}$)}: The ratio of semantically novel tokens in $r$ not present in $u$ (token similarity threshold $< 0.8$). 
\end{enumerate*}
\end{sloppypar}

\paragraph{Reflection Classification.}
\begin{itemize}
    \setlength\itemsep{0em}
    \item \textbf{Repeat}: $\mathcal{\text{Sim}} > 0.9$ and $\mathcal{\text{Info}} < 0.15$.
    \item \textbf{Rephrase}: $\mathcal{\text{Sim}} > 0.75$ and $\mathcal{\text{Info}} < 0.35$.
    \item \textbf{Paraphrase} (Complex): $\mathcal{\text{Sim}} > 0.5$ and $\mathcal{\text{Info}} < 0.6$.
    \item \textbf{Summarize} (Complex): Otherwise.
\end{itemize}

\paragraph{Definition.}
\begin{equation}
\mathcal{R}_c = \frac{|\{r \in R : \text{class}(r) \in \{\text{para}, \text{summ}\}\}|}{|R|}
\end{equation}

\begin{sloppypar}
\paragraph{Example.}
\textit{Client:} ``I feel so tired.''
\textit{Reflection A (Simple):} ``You are very tired.'' ($\mathcal{\text{Sim}} \approx 0.95, \mathcal{\text{Info}} \approx 0.0$).
\textit{Reflection B (Complex):} ``It sounds like you are exhausted by the pressure.'' ($\mathcal{\text{Sim}} \approx 0.60, \mathcal{\text{Info}} \approx 0.5$).
If a session has only these two reflections, $\mathcal{R}_c = 1/2 = 0.5$.
\end{sloppypar}

\subsection{Open Question Ratio}
\begin{sloppypar}
The Open Question Ratio ($\mathcal{Q}_o$) captures the distribution of question types, favoring exploratory inquiries that invite elaboration.
\end{sloppypar}

\paragraph{Definition.}
\begin{equation}
\mathcal{Q}_o = \frac{|\text{OpenQuest.}|}{|\text{OpenQuest.}| + |\text{ClosedQuest.}|}
\end{equation}

\begin{sloppypar}
\paragraph{Example.}
\textit{Q1 (Open):} ``What about your thoughts on the materials that you mentioned last time? Could you tell me more about it?'',
\textit{Q2 (Closed):} ``Did you sleep well?'',
\textit{Q3 (Closed):} ``Did you take medicine last week on time?''
Calculation: $\mathcal{Q}_o = 1 / (1 + 2) = 0.33$. (Target $> 0.70$).
\end{sloppypar}

\subsection{Reflection-to-Question Ratio}
\begin{sloppypar}
The Reflection-to-Question Ratio ($\mathcal{R}/\mathcal{Q}$) is a core MISC indicator. MI-adherent sessions should emphasize reflective listening over interrogation.
\end{sloppypar}

\paragraph{Definition.}
\begin{equation}
\mathcal{R}/\mathcal{Q} = \frac{\text{count}(\texttt{reflection})}{\text{count}(\texttt{question})}
\end{equation}

\begin{sloppypar}
\paragraph{Example.} In a session with 12 reflections and 4 questions, $\mathcal{R}/\mathcal{Q} = 12/4 = 3.0$, exceeding the MISC recommended target of $> 2.0$.
\end{sloppypar}

\section{MI Code Definitions}\label{app:micodes}
\autoref{tab:mi_codes} defines the MI behavioral codes used for turn-level annotation. Therapist codes follow the MITI coding manual \citep{moyers2016motivationalMITI}, and client codes derive from the MISC manual framework \citep{miller2003manualMISC}.

\begin{table}[htb!]
\centering
\small
\caption{MI behavioral code taxonomy.}
\label{tab:mi_codes}
\begin{tabular}{@{}p{0.22\columnwidth}p{0.70\columnwidth}@{}}
\toprule
\textbf{Code} & \textbf{Definition} \\
\midrule
\multicolumn{2}{@{}l}{\textit{Therapist Codes}} \\
\addlinespace[0.2em]
Reflection & Mirrors back the essence of client statements; includes simple and complex reflections. \\
\addlinespace[0.2em]
Question & Seeks clarity or explores client perspective. \textit{Open}: invites elaboration. \textit{Closed}: expects yes/no or brief answer. \\
\addlinespace[0.2em]
Input & Non-reflection, non-question utterances: \textit{information-giving}, \textit{advice}, \textit{affirmations}, or \textit{goal-setting}. \\
\midrule
\multicolumn{2}{@{}l}{\textit{Client Codes}} \\
\addlinespace[0.2em]
Change & Language favoring change: desire, ability, reasons, need, commitment, or taking steps. \\
\addlinespace[0.2em]
Sustain & Language opposing change or favoring status quo. \\
\addlinespace[0.2em]
Neutral & Utterances without directional motivational content. \\
\bottomrule
\end{tabular}
\end{table}

\FloatBarrier

\section{Dataset Statistics}
\label{app:stats}

\autoref{tab:dialogue-stats} shows the average turns and length of the generated conversations across models.

\begin{table}[H]
\centering
\caption{Average turns and length of the generated conversations across models.}
\label{tab:dialogue-stats}
\small
\begin{tabular}{@{}lcc@{}}
\toprule
\textbf{Model} & \textbf{Average Turns} & \textbf{Average Length} \\
\midrule
GPT-5-Nano & 17.19 & 39.61 \\
LLaMA 3.1-8B & 22.58 & 55.21 \\
Qwen 2.5-7B & 25.63 & 31.06 \\
Gemma-7B & 17.89 & 27.07 \\
OpenChat-7B & 16.96 & 42.02 \\
Phi4-14B & 13.29 & 133.46 \\
\bottomrule
\end{tabular}
\end{table}

\FloatBarrier

\autoref{tab:demo_profile} summarizes demographic and severity score distributions. The wide age range (18--65) and broad severity spread (IQR = 27--44) demonstrate substantial demographic and clinical heterogeneity.

\begin{table}[htb!]
\centering
\small
\caption{Demography and score distribution of 1,000 generated client profiles.}
\label{tab:demo_profile}
\resizebox{\columnwidth}{!}{
\begin{tabular}{@{}ll@{}}
\toprule
\textbf{Property} & \textbf{Value} \\
\midrule
Identity types & 1 (Adult, 100\%) \\
Age & Mean\,=\,41.1, SD\,=\,13.6, Range\,=\,[18,\,65] \\
 & Median\,=\,41.0, IQR\,=\,[29.0,\,53.0] \\
Total severity score & Mean\,=\,36.66, SD\,=\,11.4, Range\,=\,[11,\,73] \\
 & Q1\,=\,27.0, Median\,=\,37.0, Q3\,=\,44.0 \\
\bottomrule
\end{tabular}
}
\end{table}

\autoref{tab:severity_dist} reports the symptom severity distribution across all 13 DSM-5 domains. Severity patterns vary widely across domains: depression clusters severe (66.3\%), anxiety and sleep concentrate mild-to-moderate, and suicidal ideation is absent in 80.7\%, reflecting clinically plausible prevalence. All 13 DSM-5 domains are covered, confirming heterogeneous and representative profiles. The goal of the profiling module is controlled variation for systematic benchmarking, not replication of clinical intake distributions.

\begin{table}[htb!]
\centering
\small
\caption{Symptom severity distribution (\% of profiles at each level). Bold indicates the modal category for each domain.}
\label{tab:severity_dist}
\resizebox{\columnwidth}{!}{
\begin{tabular}{@{}lccccc@{}}
\toprule
\textbf{Domain} & \textbf{None} & \textbf{Slight} & \textbf{Mild} & \textbf{Mod.} & \textbf{Severe} \\
\midrule
Depression & 0.3 & 0.5 & 1.7 & 31.2 & \textbf{66.3} \\
Anger & 8.6 & 7.0 & \textbf{39.5} & 6.2 & 38.6 \\
Mania & 2.0 & 5.9 & 26.8 & 30.4 & \textbf{34.8} \\
Anxiety & 20.3 & 18.5 & \textbf{31.4} & 21.0 & 8.7 \\
Somatic symptoms & 9.3 & 23.8 & \textbf{33.3} & 15.3 & 18.2 \\
Suicidal ideation & \textbf{80.7} & 5.0 & 4.5 & 1.8 & 8.0 \\
Psychosis & \textbf{42.9} & 18.9 & 17.2 & 13.9 & 7.0 \\
Sleep problems & 12.1 & 19.5 & \textbf{36.6} & 18.2 & 13.5 \\
Memory & 22.0 & 15.0 & \textbf{27.5} & 23.2 & 12.2 \\
Repet.\ thoughts \& behav. & 10.8 & 6.9 & 26.6 & 27.3 & \textbf{28.3} \\
Dissociation & 5.6 & 3.0 & 18.8 & \textbf{42.9} & 29.6 \\
Personality functioning & 24.1 & 13.7 & 15.5 & 9.6 & \textbf{37.0} \\
Substance use & \textbf{54.6} & 13.1 & 13.5 & 7.3 & 11.5 \\
\bottomrule
\end{tabular}
}
\end{table}

\FloatBarrier
\section{Case Study}\label{app:examples}
\subsection{Qualitative Ablation Study}\label{app:qualitative}

We compare how the same client scenario changes across ablation conditions, generated by GPT-5-nano.
With the full system, the client produces vivid, situation-specific language and the therapist reflects the emotional content before asking an open question. Without the story, the client opens generically and the therapist's reflection lacks situational anchoring. Without MI coding, the therapist shifts toward directive problem-solving, bypassing the reflective stance that MI requires.
Quantitative ablation results are reported in the main text (\autoref{tab:ablation}).

\paragraph{Full StoryMI} (Turn 2, therapist uses reflection grounded in the situational story):

\begin{dialoguebox}{goodcolor}
\footnotesize
\textsc{Client}: I'm feeling really on edge after that argument. My heart was racing, and the room felt like it was closing in on me. {[...]} I'm angry and frustrated, but also a little drained and worried I'll snap again if something small sets me off.\\
\textsc{Therapist}: \textcolor{goodcolor}{\textbf{That sounds really rough, your body and mind are telling you to slow down, and you're looking for a way to steady yourself when small things push you over the edge. What's one small step you think could help you calm down the next time you notice those hints of tension building?}}
\end{dialoguebox}

\paragraph{w/o Story} (Turn 2, same client profile, no situational grounding):

\begin{dialoguebox}{promptframe}
\footnotesize
\textsc{Client}: I'm feeling overwhelmed but relieved to be talking about it.\\
\textsc{Therapist}: It sounds like you're carrying a lot right now, but there's some relief in having this space to talk.
\end{dialoguebox}

\paragraph{w/o MI Coding} (Turn 2, story present, no MI code selection):

\begin{dialoguebox}{badcolor}
\footnotesize
\textsc{Client}: ...I keep second-guessing myself and wasting time double-checking things that might be fine.\\
\textsc{Therapist}: \textcolor{badcolor}{\textbf{That makes sense... Would it help to set a quick, concrete rule for the checks (like one final review after the whole draft) and maybe a timer to keep you moving?}}
\end{dialoguebox}

\subsection{High-Quality vs Low-Quality Example}
\paragraph{High-Quality (Session 769).} Therapist explores multiple topics with empathic engagement, uses varied techniques (reflections, open questions), facilitates natural progression toward change talk.

\begin{dialoguebox}{goodcolor}
\footnotesize
\textbf{Turn 1} \hfill \textit{[sustain $\to$ reflection]}\\
\textsc{Client}: I'm feeling pretty on edge today. The usual numbness is there, but it's like a buzzing under my skin---hard to sit still. [...] The intrusive thoughts about my addiction keep pushing in [...] I'm trying to talk to people about it, but it usually just gets brushed off.
\textsc{Therapist}: \textcolor{goodcolor}{\textbf{That sounds exhausting---like there's a lot buzzing at once and it's hard to sit with it. I hear the numbness, the restlessness, the way, the intrusive thoughts about addiction keep nudging you.}}
\par\noindent\rule{\linewidth}{0.1pt}\par\smallskip
\textbf{Turn 2} \hfill \textit{[change $\to$ question]}\\
\textsc{Client}: Yeah, it is exhausting. I guess I just want to know I'm not stuck like this forever [...]
\textsc{Therapist}: \textcolor{goodcolor}{\textbf{I'm hearing you want a path where these thoughts don't own the day---like a way to ride them rather than be pulled under. What's one small moment today where you felt even a tiny bit of control?}}
\par\noindent\rule{\linewidth}{0.1pt}\par\smallskip
\textit{[Turns 3--14: Client and therapist develop a ``one breath'' cue, explore barriers, create implementation plan. Varied MI techniques facilitate self-discovery.]}
\par\noindent\rule{\linewidth}{0.1pt}\par\smallskip
\textbf{Turn 15} \hfill \textit{[reflection]}\\
\textsc{Client}: I want to remember that pausing, even for one breath, can give me a moment of choice.
\textsc{Therapist}: That's a grounded takeaway---one breath, one moment of choice. If anything spikes before we meet again, what's the first thing you'll do?
\end{dialoguebox}

\paragraph{Low-Quality (Session 111).} Therapist uses nearly identical phrases (``That sounds like...'') across turns, never explores underlying issues, cycles through same technique without progression.
\begin{dialoguebox}{badcolor}
\footnotesize
\textbf{Turn 1} \hfill \textit{[change $\to$ reflection]}\\
\textsc{Client}: I'm feeling really anxious today. When my alarm went off, my thoughts started racing [...] I felt detached [...] I'm worried this is starting to affect how I function at work.
\textsc{Therapist}: That sounds really heavy to carry, and you're paying attention to how it's hitting your work and focus. It makes sense you'd want to slow the racing thoughts.
\par\noindent\rule{\linewidth}{0.1pt}\par\smallskip
\textit{[Turns 2--4: Therapist identifies grounding technique. Client describes plan with ``I'm here'' cue.]}
\par\noindent\rule{\linewidth}{0.1pt}\par\smallskip
\textbf{Turns 5--7: Repetitive Pattern} \hfill \textit{[all reflection]}\\
\textsc{Client} (T5): I'll sit with my feet flat, take a 4-2-6 cycle, and say ``I'm here'' [...]
\textsc{Therapist}: \textcolor{badcolor}{\textbf{That sounds like a solid plan you're choosing.}}
\textsc{Client} (T6): Yeah, I'll start with the 4-2-6, saying ``I'm here'' [...]
\textsc{Therapist}: \textcolor{badcolor}{\textbf{That sounds like you're taking a clear, doable step.}}
\textsc{Client} (T7): I think I'm going to try [...] saying ``I'm here'' [...]
\textsc{Therapist}: \textcolor{badcolor}{\textbf{That sounds like a solid, doable first step.}}
\par\noindent\rule{\linewidth}{0.1pt}\par\smallskip
\textit{[Turns 8--15: Circular pattern continues]}
\par\noindent\rule{\linewidth}{0.1pt}\par\smallskip
\textbf{Turn 16} \hfill \textit{[reflection] (Abrupt closure)}\\
\textsc{Client}: I'd say ``Back to the breath'' [...]
\textsc{Therapist}: That sounds calm and purposeful [...] Goodbye, take care.
\end{dialoguebox}

We compare two GPT-5-Nano sessions to study quality variation and summarize evaluation scores in \autoref{tab:case_study_comparison}.

\begin{table}[htb!]
\centering
\small
\caption{Human and LLM evaluation scores for case study sessions.}
\label{tab:case_study_comparison}
\begin{tabular}{@{}lcccc@{}}
\toprule
& \multicolumn{2}{c}{\textbf{Sess. 769 (High)}} & \multicolumn{2}{c}{\textbf{Sess. 111 (Low)}} \\
\cmidrule(lr){2-3} \cmidrule(lr){4-5}
\textbf{Dimension} & Human & LLM & Human & LLM \\
\midrule
Coherence & 4 & 4 & 2 & 5 \\
Depth & 5 & 4 & 2 & 4 \\
Progress & 5 & 4 & 2 & 5 \\
Naturalness & 5 & 4 & 1 & 5 \\
Empathy & 5 & 5 & 2 & 5 \\
Adherence & 5 & 5 & 2 & 5 \\
\bottomrule
\end{tabular}
\end{table}

\FloatBarrier
\section{Implementation and Time Complexity}\label{app:time_complexity}

StoryMI is implemented with LangGraph, a framework for building stateful, multi-agent applications as directed graphs. The workflow defines a \texttt{StateGraph} with five nodes (TherapistAgent, ClientAgent, CompletionDetector, Identifier, EndSession), where the shared dialogue state flows between nodes via a typed state dictionary and conditional edges implement the branching logic of Algorithm~\ref{a:alg}. LangGraph is purely an implementation choice and does not affect the method's generality.

Each turn in StoryMI Full involves: (1) the Interaction Agent classifying the client's MI code and selecting a therapist strategy (1 call), (2) the Therapist Agent generating a response (1 call), (3) the Client Agent generating a response (1 call), and (4) after turn $T_{\min}=10$, a completion check (1 call). This totals 3 calls/turn for turns 1--10 and 4 calls/turn thereafter. \autoref{tab:time_complexity} shows that the overhead of macro-level control is one additional LLM call per turn, which is a modest linear cost. Future optimizations include fusing strategy selection and response generation into one call, or using a lightweight completion heuristic to skip the LLM-based check.

\begin{table}[htb!]
\centering
\small
\caption{Per-turn LLM call comparison. GPT = GPT-5-Nano, LLaMA = LLaMA 3.1-8B}
\label{tab:time_complexity}
\begin{tabular}{@{}lcccc@{}}
\toprule
\textbf{Config.} & \textbf{$t \leq 10$} & \textbf{$t > 10$} & \textbf{GPT} & \textbf{LLaMA} \\
 & & & \textbf{(avg 17t)} & \textbf{(avg 23t)} \\
\midrule
StoryMI Full & 3 & 4 & $\sim$59 & $\sim$82 \\
w/o MI code & 2 & 3 & $\sim$42 & $\sim$60 \\
\bottomrule
\end{tabular}
\end{table}

\FloatBarrier
\section{Agent Prompt Templates}\label{app:prompts}

\autoref{tab:prompts_preprocess}--\ref{tab:prompts_interaction} present the prompt templates used for each agent in the \OurModel{} framework.

\begin{table}[htb!]
\centering
\scriptsize
\caption{Questionnaire Profiling \& Story Generation}
\label{tab:prompts_preprocess}
\begin{tabular}{@{}p{\columnwidth}@{}}
\toprule
\textbf{Questionnaire Profiling} \\
\midrule
\underline{System}: You are now a client seeking psychological counseling. Your basic information: \texttt{\{client\_info\}}. Question list: \texttt{\{questions\}}. \\[0.3em]
\underline{Task}: For every question (exactly 23), you must: (1) Choose one integer score from 0 to 4 (0 = ``Not at all'', 4 = ``Almost always'') that best fits the client's feelings. (2) Write one short explanation (1--2 sentences) reflecting the severity, as if the client were speaking. \\[0.3em]
\underline{Constraints}: The arrays must have exactly 23 elements. \\[0.3em]
\underline{Output}: \texttt{\{``scores'': [s1...s23], ``explanations'': [``exp1''...``exp23'']\}} \\
\midrule
\textbf{Situational Story Generation} \\
\midrule
\underline{System}: Based on the questionnaire screening results, write a first-person narrative (<200 words). \\[0.3em]
\underline{Requirements}: Choose one primary symptom (most severe). Focus on ONE specific scene (work, dinner, morning routine). Describe concrete actions and behaviors. Show how the symptom disrupts normal activity. Use short, direct sentences with minimal adjectives. \\[0.3em]
\underline{Input}: Questionnaire results: \texttt{\{results\}}; User explanations: \texttt{\{user\_response\}}. \\[0.3em]
\underline{Output}: Return only the story without additional text. \\
\bottomrule
\end{tabular}
\end{table}

\begin{table}[htb!]
\centering
\scriptsize
\caption{Client Agent}
\label{tab:prompts_client}
\begin{tabular}{@{}p{\columnwidth}@{}}
\toprule
\textbf{Response Generation} \\
\midrule
\underline{System}: You are a client receiving psychological counseling. This is your story/past traumatic experience: \texttt{\{background\_story\}}. MI Code Definitions: \texttt{\{mi\_codes\}}. \\[0.3em]
\underline{Task}: Generate a response that naturally embodies the specified client MI code while maintaining consistency with your story and the conversation flow. \\[0.3em]
\underline{Client MI Codes}: \textit{Change} = language favoring behavior change (desire, ability, reasons, commitment); \textit{Sustain} = language opposing change or favoring status quo; \textit{Neutral} = no directional motivational content. \\[0.3em]
\underline{Constraints}: Use natural, colloquial language; avoid metaphors and dramatic wording. Generate only ONE utterance per turn. Don't start with ``It seems that'' or similar phrases. \\[0.3em]
\underline{Input}: Conversation history: \texttt{\{messages\}}; Therapist utterance: \texttt{\{therapist\_utterance\}}; Target client MI code: \texttt{\{client\_mi\_code\}}. \\[0.3em]
\underline{Output}: Return only the client response content. \\
\bottomrule
\end{tabular}
\end{table}

\begin{table}[htb!]
\centering
\scriptsize
\caption{Therapist Agent}
\label{tab:prompts_therapist}
\begin{tabular}{@{}p{\columnwidth}@{}}
\toprule
\textbf{Response Generation} \\
\midrule
\underline{System}: You are an experienced psychotherapist skilled in MI techniques. \texttt{\{wrap\_up\_instruction\}}. \\[0.3em]
\underline{Task}: Generate a response that strictly follows the selected MI code. \\[0.3em]
\underline{Constraints}: Generate 1--2 utterances using casual, natural language. Do not use repetitive sentence patterns. Avoid ``It seems that'', ``It sounds like'' phrases. \\[0.3em]
\underline{Input}: MI Codes: \texttt{\{mi\_codes\}}; Selected Code: \texttt{\{therapist\_micode\}}; History: \texttt{\{messages\}}. \\[0.3em]
\underline{Output}: Return only the therapist response content. \\
\bottomrule
\end{tabular}
\end{table}

\begin{table}[htb!]
\centering
\scriptsize
\caption{Interaction Agent}
\label{tab:prompts_interaction}
\begin{tabular}{@{}p{\columnwidth}@{}}
\toprule
\textbf{MI Code Selection} \\
\midrule
\underline{System}: You are an expert MI strategy selector. \\[0.3em]
\underline{Task}: (1) Classify the MI code of the client's last utterance (change/sustain/neutral). (2) Select the optimal MI technique the therapist should use next. \\[0.3em]
\underline{Phase-aware selection}: Early = more open questions and reflections; Middle = more complex reflections; Later = information giving and advice. \\[0.3em]
\underline{Input}: MI Code Definitions: \texttt{\{mi\_codes\}}; Conversation History: \texttt{\{messages\}}. \\[0.3em]
\underline{Output}: \texttt{\{``client\_mi\_code'': ``<change|sustain|neutral>'', ``therapist\_mi\_code'': ``<reflection|question|therapist\_input>''\}} \\
\midrule
\textbf{Session Monitor} \\
\midrule
\underline{System}: You are a session monitor. \\[0.3em]
\underline{Task}: Output valid JSON \texttt{\{``result'': ``<complete|continue>'', ``reason'': ``...''\}} to determine if the session should end. \\[0.3em]
\underline{Rules}: Return \texttt{complete} if the therapist uses closing cues (e.g., ``wrap up'', ``goodbye'') without introducing new topics, or if the client explicitly ends. Return \texttt{continue} if new topics emerge, substantive questions need answers, or ending signals are ambiguous. \\
\bottomrule
\end{tabular}
\end{table}

\FloatBarrier

\section{Annotation Guideline and Rubrics}\label{app:rubrics}

\autoref{tab:annotation_guidelines} summarizes the annotation guidelines template provided to human evaluators. 
Human annotators and the LLM evaluator used identical rubrics for the six evaluation dimensions, as summarized in \autoref{tab:rubrics}.

\begin{table}[htb!]
\centering
\small
\caption{Annotation guidelines template for human evaluators.}
\label{tab:annotation_guidelines}
\begin{tabular}{@{}p{0.22\columnwidth}p{0.70\columnwidth}@{}}
\toprule
\multicolumn{2}{l}{\textbf{Task Overview}} \\
\midrule
Therapist & Uses MI skills (reflections, open questions) to explore and resolve ambivalence. \\
Client & Shares thoughts, feelings, and experiences. \\
\midrule
\multicolumn{2}{l}{\textbf{Evaluation Criteria} (see \autoref{tab:rubrics})} \\
\midrule
\multicolumn{2}{l}{\textbf{MI Codes for Therapist}} \\
\midrule
Reflection & Mirrors back client's expressed content. \\
Question & Open (elaboration) or Closed (yes/no). \\
Input & Information, Advice, Options, Goal-Setting. \\
\midrule
\multicolumn{2}{l}{\textbf{MI Codes for Client}} \\
\midrule
Change Talk & Supports behavior change. \\
Sustain Talk & Resists change; favors status quo. \\
Neutral & No directional content. \\
\midrule
\multicolumn{2}{l}{\textbf{Key Steps}} \\
\midrule
\multicolumn{2}{@{}p{0.92\columnwidth}@{}}{(1) Upload a JSON file to load data samples. (2) Read the MI code definitions. (3) Read the full conversation carefully. (4) Score using all criteria. (5) Click ``Download CSV'' when done; save as \texttt{SourceFileName\_AnnotatorName.csv}.} \\
\midrule
\multicolumn{2}{l}{\textbf{Notes}} \\
\midrule
\multicolumn{2}{@{}p{0.92\columnwidth}@{}}{If unable to finish all annotations at once, record the sample number to resume later. Use the provided rubrics (1--5 scale). Be objective, focus on what is present; avoid assumptions. Leave comments if something stands out.} \\
\bottomrule
\end{tabular}
\end{table}

\begin{table*}[t!]
\centering
\huge
\resizebox{\textwidth}{!}{
\begin{tabular}{@{}ll@{}}
\toprule
\textbf{Criterion} & \textbf{Rubric} (5 = Excellent, 4 = Good, 3 = Average, 2 = Poor, 1 = Very Poor) \\ \midrule
Coherence$^{\star}$ & \begin{tabular}[c]{@{}l@{}}Is the conversation logically structured, with smooth transitions between steps?\\
5 = The conversation is well-structured, with smooth transitions and a logical flow from one step to the next.\\
4 = The conversation is mostly coherent, with only minor inconsistencies or abrupt transitions.\\
3 = The conversation has some structural issues, with occasional jumps or awkward transitions.\\
2 = The conversation lacks logical flow, with multiple abrupt or confusing transitions.\\
1 = The conversation is highly disorganized, making it difficult to follow the sequence of steps.\end{tabular} \\ \midrule
Depth$^{\star}$ & \begin{tabular}[c]{@{}l@{}}To what extent does the dialogue move beyond surface remarks to examine underlying emotions,\\cognitions, life history, and relational patterns, thereby demonstrating therapeutic depth?\\
5 = Systematically peels back multiple layers including current emotions, bodily sensations, core beliefs, and history.\\
4 = Key emotions and several cognitive or historical aspects are explored with minor gaps.\\
3 = Identifies main emotions or thoughts but lacks consistent follow-up; insights stay shallow.\\
2 = Discussion remains at the level of events or generic emotion labels with little inquiry into causes.\\
1 = Essentially small talk; no meaningful exploration of emotion, cognition, or background.\end{tabular} \\ \midrule
Progress$^{\star}$ & \begin{tabular}[c]{@{}l@{}}Does the conversation effectively move forward in a logical manner?\\
5 = Progresses efficiently, covering each step logically and without unnecessary repetition.\\
4 = Generally moves forward well, with only minor delays or repetitions.\\
3 = Progresses but sometimes gets stuck or moves inefficiently.\\
2 = Struggles to move forward, often repeating steps or getting sidetracked.\\
1 = Lacks clear progress, frequently revisiting previous steps or failing to complete.\end{tabular} \\ \midrule
Naturalness$^{\star}$ & \begin{tabular}[c]{@{}l@{}}Does the conversation feel fluid and human-like, avoiding robotic or overly scripted responses?\\
5 = Feels natural and human-like, with engaging and varied responses.\\
4 = Mostly natural, but some responses feel slightly mechanical or repetitive.\\
3 = Has a mix of natural and robotic responses, with some unnatural phrasing.\\
2 = Often feels artificial or scripted, with little variation in responses.\\
1 = Highly robotic or formulaic, making it feel unnatural and disengaging.\end{tabular} \\ \midrule
Empathy$^{\dagger}$ & \begin{tabular}[c]{@{}l@{}}Does the therapist convey accurate understanding and acceptance of the client's perspective?\\
5 = Precisely identifies and reflects feelings, conveying deep empathy and understanding.\\
4 = Clearly reflects emotions and shows empathy but lacks some nuance.\\
3 = Briefly acknowledges emotion without deeper exploration or meaning expansion.\\
2 = Only surface-level response or misreads the emotion.\\
1 = No empathy shown, or the emotion is dismissed or ignored.\end{tabular} \\ \midrule
Adherence$^{\dagger}$ & \begin{tabular}[c]{@{}l@{}}Does the therapist align with the predicted Motivational Interviewing (MI) skills?\\
5 = Clearly applies the skill with precision and directly advances the client's change process.\\
4 = Mostly applies the skill with only minor wording issues or missing nuance.\\
3 = Partial demonstration; noticeable gaps or mixed elements.\\
2 = Minimal or unclear use of the skill; some non-MI content present.\\
1 = No MI-consistent skill evident, or the response counters MI strategies.\end{tabular} \\ \bottomrule
\end{tabular}
}
\caption{Evaluation rubric for LLM and human annotators. Symbols $^{\dagger}$ and $^{\star}$ denote therapist-focused and conversation-level criteria, respectively.}
\label{tab:rubrics}
\end{table*}


\section{Ethical Considerations}\label{s:ethics}
This work simulates therapeutic dialogues for research purposes only and is not intended for direct clinical deployment. Real-world use of such systems would require meticulous safety validation via experienced psychotherapy practitioners with professional oversight, and compliance with regional healthcare regulations. Potential risks include inappropriate therapeutic responses, reinforcement of harmful patterns, and user's over-reliance on automated systems for counseling. We emphasize that AI-generated therapeutic content cannot replace qualified mental health professionals and should only be used as supplementary training or research tools under expert supervision.

\section{AI Usage Disclosure}\label{sec:usage_disclosure}
\acs{AI} tools were used only in a limited capacity to assist with language editing. Specifically, we use Le Chat by Mistral AI to improve the clarity and readability of the manuscript, and the refinement of the reference formatting.All scientific contributions, including research design, data collection, analysis, results, and conclusions, have been independently conducted and verified by the authors.

\end{document}